\documentclass[10pt,twocolumn,letterpaper]{article}

\usepackage[pagenumbers]{cvpr}   

\usepackage{booktabs}
\usepackage{tabularx}
\usepackage{array}
\usepackage{makecell}
\usepackage{amsmath}

\definecolor{cvprblue}{rgb}{0.21,0.49,0.74}
\usepackage[pagebackref,breaklinks,colorlinks,allcolors=cvprblue]{hyperref}

\def\paperID{*****}
\def\confName{CVPR}
\def\confYear{2026}

\newcolumntype{L}[1]{>{\raggedright\arraybackslash}p{#1}}
\newcolumntype{Y}{>{\raggedright\arraybackslash}X}

\title{Dual-Branch Remote Sensing Infrared Image Super-Resolution}

\author{
Xining Ge$^{1}$ \quad
Gengjia Chang$^{2}$ \quad
Weijun Yuan$^{3}$ \quad
Zhan Li$^{3}$ \quad
Zhanglu Chen$^{3}$\\
Boyang Yao$^{3}$ \quad
Yihang Chen$^{3}$ \quad
Yifan Deng$^{3}$ \quad
Shuhong Liu$^{4,\dag}$\\[0.6em]
$^{1}$Hangzhou Dianzi University \quad
$^{2}$Hefei University of Technology \quad
$^{3}$Jinan University\\
$^{4}$The University of Tokyo
}

\begin{document}
\maketitle

\begin{abstract}
Remote sensing infrared image super-resolution aims to recover sharper thermal observations from low-resolution inputs while preserving target contours, scene layout, and radiometric stability. Unlike visible-image super-resolution, thermal imagery is weakly textured and more sensitive to unstable local sharpening, which makes complementary local and global modeling especially important. This paper presents our solution to the NTIRE 2026 Infrared Image Super-Resolution Challenge, a dual-branch system that combines a HAT-L branch and a MambaIRv2-L branch. The inference pipeline applies test-time local conversion on HAT, eight-way self-ensemble on MambaIRv2, and fixed equal-weight image-space fusion. We report both the official challenge score and a reproducible evaluation on 12 synthetic times-four thermal samples derived from Caltech Aerial RGB-Thermal, on which the fused output outperforms either single branch in PSNR, SSIM, and the overall Score. The results suggest that infrared super-resolution benefits from explicit complementarity between locally strong transformer restoration and globally stable state-space modeling.
\end{abstract}

\section{Introduction}
Image super-resolution is a fundamental task in low-level vision that underpins a wide range of downstream applications, including autonomous driving~\cite{lisgs2025,liumg2025,zhou2024mod}, VR/AR~\cite{lidense2025}, 3D reconstruction under adverse conditions~\cite{liu2025i2nerf,cui2026unifying,liu2026ntire}, and visual restoration and scene understanding in low-quality images~\cite{liuderain2025,liu2025realx3d,liudenoise2026,ge2026clip}. Among its many variants, remote sensing image super-resolution plays a particularly important role in aerial observation, thermal monitoring, and downstream interpretation, since spatial resolution directly affects target visibility, contour clarity, and region-level reasoning~\cite{wang2022comprehensive,wang2022review,aleissaee2023transformers}. In thermal imagery, this issue is especially pronounced. Object interiors are often smooth, gradients are weak, and a small change in boundary quality may matter more than aggressive texture generation~\cite{rivadeneira2021thermal,rivadeneira2024thermal,zhang2021infrared,patel2021thermisrnet}. Infrared super-resolution therefore requires not only perceptual enhancement but also faithful structural recovery.

Recent super-resolution backbones include transformer-based architectures such as SwinIR \cite{liang2021swinir}, Uformer \cite{wang2022uformer}, Restormer \cite{zamir2022restormer}, and HAT \cite{chen2023activating}, as well as state-space models such as MambaIR \cite{guo2024mambair} and MambaIRv2~\cite{guo2025mambairv2}. These architectures offer different tradeoffs between local detail modeling, long-range information propagation, and computational efficiency. For infrared imagery, such differences are not merely architectural preferences. A model that sharpens locally may help recover thermal contours, while a model with stronger global propagation may better preserve scene-wide stability and suppress unstable hot spots. This observation motivates a design that explicitly leverages both kinds of inductive bias rather than committing to a single backbone family.

In this paper, we present our solution to the NTIRE 2026 Infrared Image Super-Resolution Challenge \cite{liu2026rsirsr}, a dual-branch system that combines a HAT-L branch and a MambaIRv2-L branch. The inference pipeline applies test-time local conversion (TLC) on the HAT branch, eight-way self-ensemble on the MambaIRv2 branch, and fixed equal-weight image-space fusion to merge the two outputs. The two branches are deliberately complementary. HAT-L with TLC provides locally strong transformer restoration that recovers fine boundaries, while MambaIRv2-L with self-ensemble provides globally stable state-space modeling that maintains radiometric consistency across the scene. We evaluate the proposed system on the official NTIRE 2026 hidden test set and additionally on a reproducible evaluation set of 12 synthetic times-four thermal samples derived from Caltech Aerial RGB-Thermal, on which the fused output outperforms either single branch in PSNR, SSIM, and the overall Score.

Our main contributions can be summarized as:
\begin{itemize}
\item We present our solution to the NTIRE 2026 Infrared Image Super-Resolution Challenge, a dual-branch framework that combines HAT-L and MambaIRv2-L with branch-specific test-time refinements to exploit their complementary local and global modeling capabilities.
\item We evaluate the proposed system on both the official challenge benchmark and a reproducible evaluation set derived from Caltech Aerial RGB-Thermal, on which the fused output consistently outperforms either single branch.
\end{itemize}

\section{Related Work}
\subsection{Image Super-Resolution Backbones}
Modern image super-resolution \cite{ren2026esr,chen2026sr4} has evolved from convolutional reconstruction networks to transformer and state-space models. Transformer restorers such as SwinIR, Uformer, Restormer, and HAT improve reconstruction through stronger local-global interaction and more expressive feature routing~\cite{liang2021swinir,wang2022uformer,zamir2022restormer,chen2023activating}. State-space restoration models such as MambaIR and MambaIRv2 further strengthen long-range propagation while keeping the architecture lightweight and scalable~\cite{guo2024mambair,guo2025mambairv2}. In remote sensing super-resolution, related work has also explored multistage transformers, reference-based designs, lightweight fusion networks, and diffusion-style reconstruction~\cite{lei2021transformer,dong2021rrsgan,wang2023lightweight,xiao2023ediffsr,xiao2024frequency}. These developments make heterogeneous branch design a reasonable choice rather than an ad hoc engineering combination.

\subsection{Infrared and Thermal Image Super-Resolution}
Infrared super-resolution differs from visible-image SR because thermal imagery is typically sparse in texture and more dependent on stable region transitions than on dense high-frequency detail. Existing thermal and infrared SR studies have explored multiscale spatio-temporal fusion, efficient thermal SR networks, real-world infrared degradation modeling, and transformer-based infrared reconstruction~\cite{zhang2021infrared,patel2021thermisrnet,zhou2023real,chen2024modeling,liang2023dasr,qin2024lkformer}. Challenge-oriented benchmarks have further shown that thermal SR requires careful treatment of evaluation, degradation realism, and reproducibility~\cite{rivadeneira2021thermal,rivadeneira2024thermal}. More broadly, adverse-condition visual restoration has also expanded to low-light and smoke-degraded 3D reconstruction, where recent challenge reports and Gaussian-splatting-based pipelines combine multimodal priors, luminance-guided enhancement, pseudo-clean supervision, generative restoration, progressive correction, and reliability-aware modeling to recover robust geometry and appearance~\cite{liu2026ntire,zheng20263d,liu2026elog,fu2026smokegs,cao2026gensmoke,zhu2026naka,guo2026reliability,chen2026dehaze}. While these neighboring problems are not identical to infrared SR, they reinforce the broader importance of robust restoration under adverse sensing conditions. This setting motivates conservative system design in which branch complementarity may be more valuable than maximizing the sharpness of any single model in isolation.

\subsection{Test-Time Enhancement and Image Fusion}
Test-time enhancement strategies are particularly attractive when retraining is costly or infeasible. Test-time local conversion (TLC) mitigates the train-test discrepancy that arises between patch-based training and full-image inference~\cite{chu2022improving}. Recent restoration challenge reports further suggest that x8 geometric self-ensemble can provide small but consistent gains at inference time without changing the trained model~\cite{chang2026beyond}. Output-level fusion is similarly effective when individual branches exhibit complementary restoration biases. Recent training-free SR ensemble work further shows that a stable main branch and a stronger compensation branch can be fused at the image level to improve reconstruction without additional training~\cite{chang2026training}. In infrared and multimodal reconstruction, such branch-level complementarity has been shown to yield consistent gains even under simple deterministic fusion rules~\cite{xiao2022heterogeneous,wang2022multimodal}. Our system follows this design philosophy, training and running each branch independently, applying branch-specific test-time refinements to exploit their respective strengths, and combining the two outputs at the image level through fixed equal-weight fusion.

\section{Method}
\subsection{Dual-Branch Framework}
Our solution comprises two independently optimized branches, a HAT-L branch dedicated to fine-grained local contour modeling and a MambaIRv2-L branch responsible for long-range context aggregation. Both branches receive the same low-resolution infrared input and are fused only after their respective inference paths are completed,
\begin{equation}
\begin{aligned}
I_{\mathrm{HAT}} &= F_{\mathrm{HAT}}^{\mathrm{TLC}}(I_{\mathrm{LR}}), \\
I_{\mathrm{Mamba}} &= E_{\mathrm{8}}\!\left(F_{\mathrm{Mamba}}(I_{\mathrm{LR}})\right),
\end{aligned}
\end{equation}
\begin{equation}
I_{\mathrm{SR}} = 0.5 \, I_{\mathrm{HAT}} + 0.5 \, I_{\mathrm{Mamba}},
\end{equation}
where $F_{\mathrm{HAT}}^{\mathrm{TLC}}$ denotes the HAT-L branch wrapped with test-time local conversion and $E_{\mathrm{8}}$ denotes eight-way geometric self-ensemble. The framework deliberately avoids feature-level interaction or learned fusion heads. This design keeps the two branches fully decoupled, allows each to be optimized and analyzed in isolation, and ensures that the final prediction depends on a deterministic combination.

\begin{figure*}[t]
\centering
\includegraphics[width=0.93\textwidth]{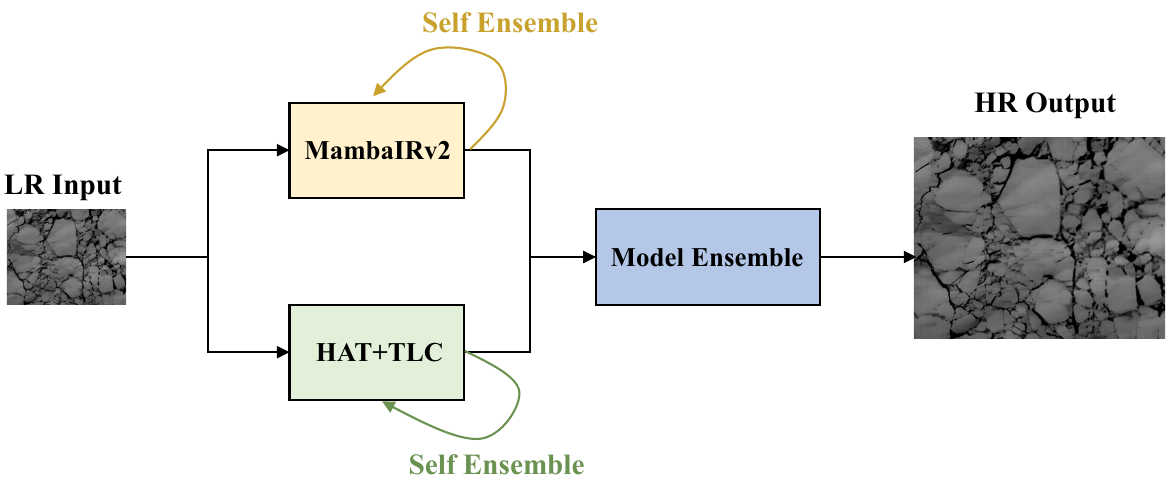}
\caption{Overview of the proposed dual-branch framework for the NTIRE 2026 remote sensing infrared super-resolution challenge. The HAT-L and MambaIRv2-L branches operate independently and their outputs are combined through fixed equal-weight image-space fusion.}
\label{fig:pipeline}
\end{figure*}
\subsection{Local and Global Branches}
The first branch is built upon HAT-L with a window size of 32. As a transformer architecture explicitly designed to activate more input pixels through hybrid attention, HAT excels at thermal edge sharpening, contour recovery, and small-structure enhancement~\cite{chen2023activating}. This capability is particularly valuable for infrared imagery, where target identity is often determined by a small number of weak boundary transitions rather than by dense high-frequency textures.
The second branch follows the official MambaIRv2 Large configuration at scale factor four~\cite{guo2025mambairv2}. By leveraging selective state-space modeling, this branch efficiently propagates information across long spatial distances and maintains large-scale structural consistency. Compared with the HAT branch, it behaves more conservatively in local detail, but it suppresses unstable local artifacts and preserves radiometric coherence at the scene level.
The two branches are therefore better understood as complementary specialists than as redundant experts. HAT-L contributes locally aggressive restoration that recovers sharp thermal contours, whereas MambaIRv2-L contributes globally stable modeling that preserves scene-wide consistency. Their predictions are merged only after branch-specific test-time refinements, which keeps the overall framework simple, interpretable, and easy to deploy.
\subsection{Test-Time Refinement}
Three test-time refinements are introduced to fully exploit the strengths of each branch. First, the HAT branch is wrapped with test-time local conversion (TLC), which restructures the inference path into a more locally consistent form and mitigates the train-test discrepancy that arises when patch-trained transformers are evaluated on full-resolution inputs~\cite{chu2022improving}. Second, the MambaIRv2 branch employs eight-way self-ensemble under horizontal flip, vertical flip, and transpose transformations, which improves prediction robustness without modifying the learned parameters. Third, the two refined outputs are merged through fixed equal-weight image-space fusion, avoiding any adaptive weighting or learned fusion module that would introduce additional parameters or training cost.
These refinements are deliberately branch-specific rather than uniformly applied. TLC is reserved for the HAT branch because the train-test mismatch it addresses is most pronounced in window-based transformer architectures, while geometric self-ensemble is applied to the MambaIRv2 branch where it most effectively stabilizes the state-space prediction. This asymmetric design reflects the complementary nature of the two branches and contributes to the overall robustness of the proposed system.

\begin{figure*}[t]
    \centering
    \includegraphics[width=\textwidth]{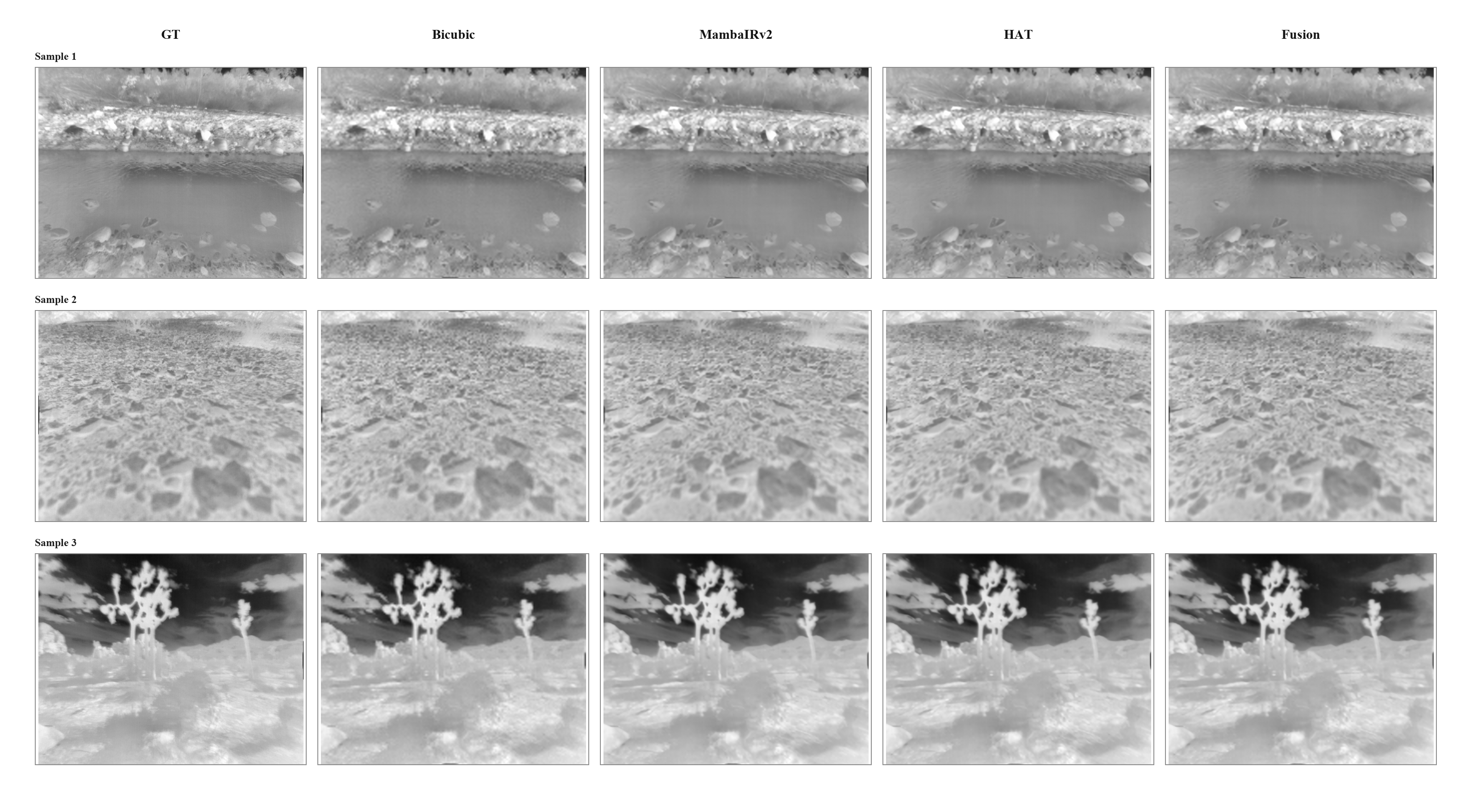}
    \caption{Qualitative comparison on representative public thermal synthetic $\times 4$ samples. Columns are GT, Bicubic, MambaIRv2, HAT, and Fusion.}
    \label{fig:qualitative}
\end{figure*}

\section{Experiments}
\paragraph{Implementation Details}
Both branches are trained on the official NTIRE 2026 development set, which contains 1019 paired infrared images. The HAT-L branch is optimized with L1 loss using AdamW for 260K iterations, while the MambaIRv2-L branch is optimized with L1 loss using Adam for 250K iterations. All training is conducted on a single NVIDIA H200 GPU. At inference, the HAT-L branch operates with a window size of 32 and is wrapped with test-time local conversion, the MambaIRv2-L branch applies eight-way geometric self-ensemble, and the two refined outputs are combined through deterministic equal-weight fusion in image space.

\paragraph{Datasets and Evaluation Protocol}
We evaluate the proposed system on two benchmarks. The first is the official NTIRE 2026 hidden test set, on which our submission achieves a final score of 54.23. The second is constructed from 12 thermal frames of the Caltech Aerial RGB-Thermal dataset, with bicubic $\times 4$ downsampling used to synthesize the low-resolution inputs. Images are modcropped and a 4-pixel border is shaved before computing metrics. The overall score is defined as $\mathrm{Score}=\mathrm{PSNR}+20\times\mathrm{SSIM}$, consistent with the official challenge protocol.

\begin{table}[t]
    \centering
    \small
    \setlength{\tabcolsep}{4pt}
    \renewcommand{\arraystretch}{1.10}
    \caption{Historical challenge submission summary.}
    \label{tab:challenge}
    \begin{tabularx}{\columnwidth}{L{0.30\columnwidth} Y}
        \toprule
        Item & Result \\
        \midrule
        Benchmark & NTIRE 2026 hidden testset \\
        Reported score & 54.23 \\
        Metric & PSNR + 20 $\times$ SSIM \\
        \bottomrule
    \end{tabularx}
\end{table}

\subsection{Quantitative Results}
Table~\ref{tab:quantitative} reports the quantitative comparison on the Caltech Aerial RGB-Thermal evaluation set. Both learned branches substantially surpass the bicubic baseline, and their fusion achieves the best overall performance across PSNR, SSIM, and the combined Score. Although the single-branch results are already highly competitive, the consistent additional gain from fusion confirms that HAT-L and MambaIRv2-L capture complementary rather than redundant aspects of infrared restoration. HAT-L favors sharper local contours while MambaIRv2-L better preserves scene-level layout, and the equal-weight combination inherits the strengths of both, producing reconstructions that are simultaneously sharper than MambaIRv2-L and more globally consistent than HAT-L.

\begin{table}[t]
\centering
\small
\setlength{\tabcolsep}{4pt}
\renewcommand{\arraystretch}{1.10}
\caption{Quantitative comparison on the Caltech Aerial RGB-Thermal evaluation set at scale $\times4$.}
\label{tab:quantitative}
\begin{tabular*}{\linewidth}{@{\extracolsep{\fill}}lccc@{}}
\toprule
Method & PSNR & SSIM & Score \\
\midrule
Bicubic   & 37.1588 & 0.9270 & 55.6982 \\
MambaIRv2 & 37.8274 & 0.9321 & 56.4690 \\
HAT       & 37.8657 & 0.9321 & 56.5070 \\
Fusion    & \textbf{37.8699} & \textbf{0.9321} & \textbf{56.5128} \\
\bottomrule
\end{tabular*}
\end{table}

\subsection{Qualitative Results}
Figure~\ref{fig:qualitative} presents representative visual comparisons on the thermal evaluation set. The HAT-L branch produces noticeably sharper local contours and recovers fine boundary transitions around small thermal targets, which is particularly important when object identity in infrared imagery is determined by a few weak edge cues rather than by dense texture. The MambaIRv2-L branch, in contrast, yields smoother and more spatially coherent reconstructions that better preserve the global radiometric distribution. The fused output retains the sharp boundaries from HAT-L while inheriting the global consistency from MambaIRv2-L, achieving a favorable balance between edge clarity and structural stability. This visual pattern reinforces the view that, for infrared imagery, faithful structural recovery is more valuable than aggressive synthesis of high-frequency texture.

\section{Conclusion}
We presented our solution to the NTIRE 2026 Infrared Image Super-Resolution Challenge, a dual-branch framework that couples a HAT-L branch with test-time local conversion and a MambaIRv2-L branch with eight-way self-ensemble, fused through fixed equal-weight image-space averaging. Experiments on both the official challenge benchmark and the Caltech Aerial RGB-Thermal evaluation set show that the fused output consistently surpasses either single branch, demonstrating that pairing locally strong transformer restoration with globally stable state-space modeling is an effective strategy for infrared super-resolution.

{\small
\bibliographystyle{plain}
\bibliography{references}
}

\end{document}